\documentclass{article}
\usepackage{ijcai26}

\usepackage{times}
\usepackage{soul}
\usepackage{url}
\usepackage[hidelinks]{hyperref}
\usepackage[utf8]{inputenc}
\usepackage[small]{caption}
\usepackage{graphicx}
\usepackage{amsmath}
\usepackage{amsthm}
\usepackage{booktabs}
\usepackage{algorithm}
\usepackage{algorithmic}
\usepackage[switch]{lineno}
\usepackage{amssymb}

\title{PILIR: Physics-Informed Local Implicit Representation}

\author{
Jianfeng Li$^1$
\and
Feng Wang$^{1,*}$
\and
Ke Tang$^2$\\
\affiliations
$^1$School of Computer Science, Wuhan University\\
$^2$School of Computer Science and Technology, Southern University of Science and Technology\\
\emails
fengwang@whu.edu.cn,
}
\begin{document}

\maketitle

\begin{abstract}
    Physics-Informed Neural Networks have become a powerful mesh-free method for solving partial differential equations, but their performance is often limited by spectral bias. Specifically, in standard MLPs used in PINNs, the global parameter coupling causes the model to prioritize learning low-frequency components, resulting in slow convergence for high-frequency details. To overcome this limitation, we introduce the Physics-Informed Local Implicit Representation (PILIR). Our approach separates the global physical domain into a discrete latent feature space and a continuous generative decoder. By using a learnable grid to encode explicit spatial locality, PILIR can capture high-frequency details locally, preventing dilution by global patterns. A generative neural operator then synthesizes these local latent features into continuous physical fields, allowing accurate reconstruction of fine-scale structures. Experiments on a range of challenging PDEs show that PILIR effectively mitigates spectral bias, thereby boosting the convergence of high-frequency details and achieving superior accuracy compared to state-of-the-art methods.
\end{abstract}

\section{Introduction}
    Physical systems are generally governed by partial differential equations (PDEs), which seldom have exact analytical solutions~\cite{willard2022integrating}. Traditional numerical approaches, such as finite element, finite volume, or finite difference methods, require significant expertise and rely heavily on carefully constructed discretizations, which can limit their flexibility. In contrast, Physics-Informed Neural Networks (PINNs)~\cite{raissi2019physics} have emerged as a promising mesh-free alternative for solving PDEs through gradient-based optimization. Benefiting from increased computational power and modern automatic differentiation tools, PINNs and related variants have been successfully applied to a diverse set of challenging PDE problems~\cite{hu2025score,li2024solving}.

    Although promising, the training process of PINNs is often plagued by slow convergence and instability, which stem from three coupled issues. First, the composite loss function introduces numerical stiffness, where the mismatch in scale between derivative terms and initial/boundary conditions creates an ill-conditioned optimization landscape~\cite{zhou2025dual,song2024loss}. This imbalance frequently leads to gradient pathologies, such as gradient masking, where dominant gradients suppress the fine updates necessary for physical fidelity~\cite{Liu2024ConFIG}. Even if these optimization issues are addressed, spectral bias remains a fundamental architectural limitation. As explained by the frequency principle~\cite{xu2025overview}, MLPs exhibit a strong inductive bias toward low-frequency components~\cite{wang2022and}, making the learning of high-frequency details exceedingly slow. Therefore, overcoming spectral bias is important for accurately capturing complex physical phenomena.

    The spectral bias in MLPs for modeling complex physical problems originates from two factors. First, from an architectural perspective, MLPs rely on global receptive fields, meaning every neuron is connected to the entire input domain. This global coupling biases the network towards capturing global and low-frequency trends, while suppressing localized high-frequency oscillations. Consequently, the network struggles to resolve fine-grained physical features without introducing artifacts in other regions. Second, from a mapping perspective, learning a direct mapping from low-dimensional coordinates to high-dimensional physical fields is difficult. Without an effective feature embedding mechanism, the network requires substantial capacity to transform these raw coordinates into a useful representation, which may reduce training efficiency and complicate convergence.

    To alleviate spectral bias, we propose PILIR, a hybrid paradigm that combines local feature extraction with continuous feature synthesis. Inspired by the local implicit image function (LIIF) in computer vision~\cite{chen2021learning}, our approach decouples the PDE solution into a discrete grid encoding and a continuous neural decoding process. This strategy directly addresses the aforementioned two bottlenecks. First, by discretizing the computational domain into a grid, we enforce explicit spatial locality, which directs the network to focus on local residuals and thereby alleviates the global parameter coupling in standard MLPs. Second, rather than querying physics from raw coordinates, we assign learnable feature vectors to grid vertices. Analogous to feature channels in image processing, these vectors serve as rich, implicit physical contexts, effectively bypassing the difficulty of low-dimensional coordinate mapping. Crucially, to bridge the gap between discrete grid vertices and continuous coordinates, a lightweight neural operator is employed for generative feature synthesis. This ensures the infinite resolution and high-order differentiability required for physics-informed modeling.

    While similar grid-based approaches have recently been adopted in PINNs to mitigate spectral bias~\cite{kang2023pixel,shishehbor2024parametric,kang2025pig}, a critical misalignment remains. Existing methods mainly rely on deterministic interpolation kernels (e.g., bilinear) to mix values, which are ill-suited for solving high-frequency PDEs. These fixed kernels enforce a convex hull constraint, limiting the solution to a weighted average of grid vertices' features, and introduce artificial numerical viscosity that smears out fine-grained structures. In contrast, we believe that the reconstruction of physical fields from discrete representations should be a generative process rather than a mechanical mixing. Accordingly, our method utilizes a neural operator to synthesize continuous values based on local contexts. This paradigm shift decouples the reconstruction quality from the grid resolution, enabling the precise capture of sub-grid details and high-frequency components.

    Various experiments demonstrate that PILIR outperforms state-of-the-art methods. By leveraging this generative decoding, our approach achieves superior accuracy in capturing high-frequency details and multi-scale variation, even at significantly lower grid resolutions.

\section{Related Works}
\subsection{Mitigation Strategies for Spectral Bias}
    Spectral bias is a well-documented phenomenon in PINNs~\cite{fridovich2022spectral}, characterized by the rapid convergence of low-frequency components alongside the stagnation of high-frequency details. Existing research has primarily explored solutions from three pathways to address this fundamental limitation.
    
    (1) Preprocessing input coordinates. These approaches introduce high-frequency inductive biases via input mapping layers, most notably Fourier feature mapping~\cite{tancik2020fourier,wang2021eigenvector,jin2024fourier}. While effective, these mappings remain globally static. Consequently, selecting an appropriate frequency bandwidth requires extensive trial and error. 
    
    (2) Improving activation functions. These methods employ multiscale activation functions rather than classical Tanh to enrich the frequency components of latent features. For instance, wavelet activations~\cite{uddin2023wavelets} enable the network to represent functions across multiple scales, thereby enhancing its ability to capture high-frequency variations. 
    
    (3) Designing novel network architectures. Recent research has investigated novel architectures like Kolmogorov-Arnold Networks (KANs) to replace traditional MLPs~\cite{wang2025kolmogorov,zhang2025physics,yang2025multi}. By employing learnable activation functions on edges, KANs offer a form of local adaptivity that mitigates the global parameter coupling inherent in MLPs. However, this flexibility comes at the cost of optimization difficulty compared to the robust MLP baseline.

\subsection{Implicit Representation and Grid Embedding}
    Implicit representations, such as LIIF and its successors~\cite{gao2023implicit,saragadam2023wire}, have established a promising paradigm in computer vision for continuous signal generation from discrete pixel representations. By employing a coordinate-based neural network as an image contexts decoder, these methods achieve resolution-independent reconstruction, effectively decoupling the learned representations from any fixed grid. 
    
    This core concept of generating continuity from discrete elements finds a direct counterpart in scientific computing through grid embedding techniques~\cite{muller2022instant,fridovich2023k}. Adapted for PINNs, these methods discretize the domain into a grid of learnable features, providing a local prior that effectively mitigates spectral bias and enhances the learning of high-frequency details~\cite{wang2024neural,gao2024physics}. The explicit grid structure also simplifies the enforcement of domain-specific constraints, such as periodic boundary conditions. 
    However, these methods rely on deterministic interpolation kernels (e.g., bilinear) to query continuous fields from discrete feature grids. This fixed mapping can restrict expressive power and reconstruction fidelity. Motivated by the structural analogy between image pixels and physical feature grids, we leverage the grid as a structured latent representation but replace deterministic interpolation with a learned neural decoder. This approach aims to synthesize continuous physical fields with higher fidelity, moving beyond the limitations of predefined interpolation schemes.

\begin{figure*}[htbp]
\centering
\includegraphics[width=0.9\linewidth]{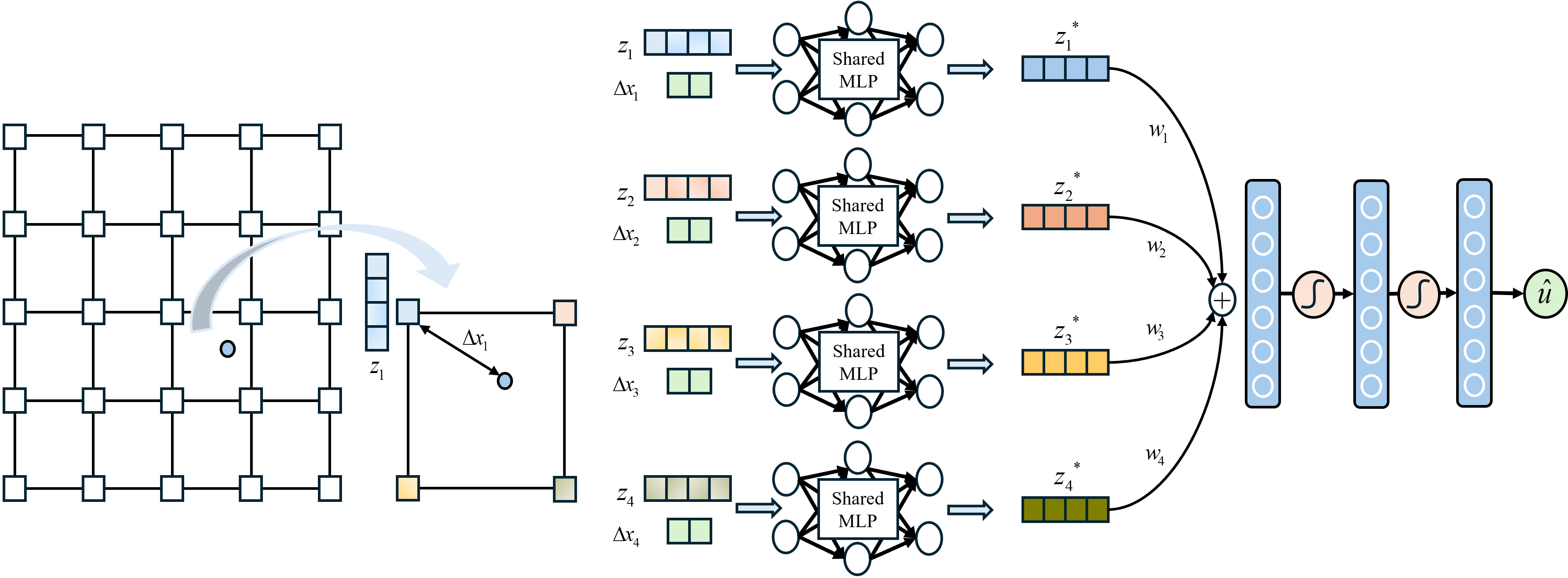}
\caption{The workflow of PILIR.}
\label{fig1}
\end{figure*}

\section{Methods}
    PILIR reformulates PDE solving as a discrete-to-continuous reconstruction process, structured within an encoder-decoder paradigm. The architecture consists of three distinct stages: (1) Domain Discrete Encoding, where the computational domain is discretized into a learnable feature grid to capture local latent contexts; (2) Continuous Feature Synthesis, where a coordinate-based neural network fuses these local discrete contexts to generate continuous feature representations for arbitrary query coordinates; and (3) Physical Decoding, which maps the synthesized features to the target physical quantities, thereby recovering the continuous physical fields with high fidelity. Figure \ref{fig1} illustrates the workflow of PILIR.
\subsection{Preliminary: Physics-Informed Loss}
    Let us begin by reviewing the content of physics-informed loss. Consider a general formulation of time-dependent PDEs with well-defined initial and boundary functions as follows:
    \begin{align}
        &\mathcal{F}_{x,t}[u](x,t)=f(x,t), &&x\in \Omega, t\in [0,T] \\
        &u(x,0)=h(x), &&x\in \Omega \\ 
        &\mathcal{B}_{x,t}[u](x,t)=g(x,t), &&x\in \partial \Omega, t\in [0,T]
    \end{align}
    where $u$ is the exact solution of corresponding PDE, $\mathcal{F}_{x,t}$ is a differential operator in terms of the spatial coordinates $x$ and the temporal coordinate $t$, $\mathcal{B}_{x,t}$ is a boundary operator, and the right-hand side functions $f,h,g$ are all known. Physics-informed neural networks try to find an approximation solution $u_{\theta}(x,t)=\mathcal{N}_{\theta}(x,t)$ using a neural network $\mathcal{N}$ with trainable parameters $\theta$ by minimizing:
    \begin{equation}
        \label{eq4}
        \begin{aligned}
            \mathcal{L}(\theta)=&\frac{\lambda_{r}}{N_{r}}\sum_{i}^{N_{r}}{(\mathcal{F}_{x_i,t_i}[u_{\theta}](x_i,t_i)-f(x_i,t_i))^2} \\
            &+\frac{\lambda_{ic}}{N_{ic}}\sum_{j}^{N_{ic}}{(u_{\theta}(x_j,0)-h(x_j,0))^2} \\
            &+\frac{\lambda_{bc}}{N_{bc}}\sum_{k}^{N_{bc}}{(\mathcal{B}_{x_k,t_k}[u_{\theta}](x_k,t_k)-g(x_k,t_k))^2},
        \end{aligned}
    \end{equation}
    where $\lambda_r,\lambda_{ic},\lambda_{bc}$ are weighting factors for PDE residual, initial condition, and boundary condition loss, respectively, and $N_r,N_{ic},N_{bc}$ denote the corresponding numbers of sample points. The derivatives of $u_{\theta}$ w.r.t. input coordinates are calculated via automatic differentiation.

\subsection{Domain Discrete Encoding}
    Consider a computational domain $\Omega \subset \mathbb{R}^d$ for a given PDE problem, where the time dimension is regarded as a special spatial dimension for convenience. To focus the network on local information, we partition $\Omega$ into a set of disjoint local regions using a structured grid. Then, a learnable discrete feature space $\mathcal{Z} \in \mathbb{R}^{N_1 \times \dots \times N_d \times C}$ can be defined, where $N_i$ denotes the number of grid vertices along the $i$-th dimension, and $C$ represents the channel dimension of the latent feature vector. Unlike previous methods that employed multi-resolution grid encoding to learn features at different scales, our approach utilizes a single-resolution grid solely for acquiring local information. Fine details are learned through subsequent Continuous Feature Synthesis, thereby reducing storage requirements for grid parameters.

    Then, for any continuous query coordinate $\mathbf{x} \in \Omega$, we first normalize it to the range $[0, 1]^d$. The spatial locality is enforced by locating the voxel that encloses $\mathbf{x}$. To elaborate, considering a two-dimensional case, the coordinate of the top-left (or north-west) grid vertex is regarded as an anchor index, determined by the floor operation, i.e., $\mathbf{x}_{anchor}=\lfloor \mathbf{x} \cdot \mathbf{N}\rfloor$, where $\mathbf{N}=(N_1,N_2,\cdots ,N_d)$ represents the grid resolution. 

\subsection{Continuous Feature Synthesis}
    After Domain Discrete Encoding, we need to reconstruct a continuous feature representation for any arbitrary query coordinate. Unlike standard interpolation that employs mechanical numerical mixing, our approach treats feature synthesis as a local generative process. 
    
    For a given query point $\mathbf{x}$, we locate its enclosing grid voxel and retrieve the local context based on the anchor index $\mathbf{x}_{anchor}$. Let $\mathcal{N}(\mathbf{x}) = \{v_1, \dots, v_{2^d}\}$ denote the set of $2^d$ corner vertices of the enclosing voxel. We extract the corresponding latent feature vectors $\mathcal{Z}_{\mathbf{x}} = \{\mathbf{z}_i\}_{i=1}^{2^d}$ from the grid, where $\mathbf{z}_i \in \mathbb{R}^C$ denotes the latent feature vector stored in $v_i$. Simultaneously, to ensure translation invariance and facilitate the learning of local spatial frequencies, we compute the relative coordinate offset for each vertex: 
    \begin{equation}
        \Delta \mathbf{x}_i=\mathbf{x}-\mathbf{x}_i, \forall i \in \{1, \cdots ,2^d\},
    \end{equation}
    where $\mathbf{x}_i$ represents the coordinate of the vertex $v_i$.

    Then, a shared neural decoding operator $f_{\theta_s}: \mathbb{R}^C \times \mathbb{R}^d \to \mathbb{R}^H$, parameterized as a MLP, is employed to learn a synthesized high-dimensional feature contribution based on both the content of the vertex (local latent contexts) and the geometric relationship (relative offset). For example, the predicted feature contribution from the $i$-th neighbor is computed as:
    \begin{equation}
    \label{eq6}
        \mathbf{h}_i(\mathbf{x})=f_{\theta_s}(\mathbf{z}_i, \Delta \mathbf{x}_i).
    \end{equation}
    Here, $f_{\theta_s}$ acts as a local implicit function that decodes the discrete local latent contexts into the continuous feature space.

    Finally, we aggregate the feature contributions from all neighbours $\{\mathbf{h}_i\}_{i=1, \cdots, 2^d}$ via a spatial weighting mechanism to obtain the complete feature representation at $\mathbf{x}$. This is because when $\mathbf{x}$ is located near both sides of a voxel boundary, the network will utilize different vertex contributions to the feature representation, resulting in a feature confusion. Therefore, to guarantee continuity across boundaries, the synthesized feature representation $\mathbf{h}(\mathbf{x})$ is formulated as:
    \begin{equation}
    \label{eq7}
        \mathbf{h}(\mathbf{x}) = \sum_{i}^{2^d}w_i(\mathbf{x})\cdot \mathbf{h}_i=\sum_{i}^{2^d}w_i(\mathbf{x})\cdot f_{\theta_s}(\mathbf{z}_i, \Delta \mathbf{x}_i),
    \end{equation}
    where $w_i(\mathbf{x})$ represents the normalized volume-based weight corresponding to vertex $v_i$, e.g., bilinear. For PDEs that require higher-order derivatives, we utilize cosine re-weighting proposed by~\cite{kang2023pixel} to guarantee $C^{\infty}$ continuity, written as:
    \begin{equation}
        w_i^{'}(\mathbf{x}) = \frac{1}{2}(1-cos(\pi \cdot w_i(\mathbf{x}))).
    \end{equation}

    While both PILIR and standard interpolation, typically defined as $\mathbf{h}_{std}(\mathbf{x}) = \sum_{i} w_i \mathbf{z}_i$, utilize spatial weighting $w_i$ to ensure smooth transitions across voxel boundaries, a fundamental distinction lies in the operand. In our approach, the weighting is applied to the non-linear projections of the local latent contexts ($f_{\theta_s}(\mathbf{z}_i, \Delta \mathbf{x}_i)$) rather than the raw contexts themselves. Here, the grid features $\mathbf{z}_i$ function not as scalar physical intensities, but as generative parameters. Consequently, high-frequency details are synthesized by the neural operator $f_{\theta_s}$ within the voxel based on these parameters, allowing the reconstructed field to exhibit complex non-linear variations and transcend the rigid convex hull constraint inherent in traditional linear mixing at low resolution.

\begin{algorithm}[H]
\caption{Main Process of PILIR}
\label{alg:method}
\textbf{Input}: Sample points inside the domain $\mathcal{D}_f \subset \Omega$, Boundary/Initial points $\mathcal{D}_b \subset \partial \Omega$, Grid resolution $\mathbf{N}$. \\
\textbf{Output}: Optimized discrete latent features $\mathcal{Z}$, Continuous feature synthesis network parameters $\theta_s$, Physical decoding head parameters $\psi$.
\begin{algorithmic}[1]
\STATE \textbf{Initialization:} 
\STATE Initialize discrete latent features $\mathcal{Z} \in \mathbb{R}^{N_1 \times \dots \times N_d \times C}$ with uniform distribution.
\STATE Initialize neural networks $f_{\theta_s}$ and $\phi_\psi$ with Xavier normal initialization.
\WHILE{not converged}
    \STATE Random sample a batch of coordinate points $\mathbf{x}$ from $\mathcal{D}_f$ and $\mathcal{D}_b$, respectively.
    \FOR{each query point $\mathbf{x}$ in batch}
        \STATE Retrieve $2^d$ neighbor latent vectors $\{\mathbf{z}_i\}_{i=1}^{2^d}$ from $\mathcal{Z}$ according to $\mathbf{x}_{anchor} = \lfloor \mathbf{x} \cdot \mathbf{N} \rfloor$.
        \FOR{each {$i \in \{1,\cdots,2^d\}$}}
            \STATE Compute relative offsets $\Delta \mathbf{x}_i = \mathbf{x} - \mathbf{x}_i$ and spatial weights $w_i(\mathbf{x})$.
            \STATE Synthesize local feature contribution from $i$-th neighbor $\mathbf{h}_i(\mathbf{x})$ via the neural operator, using Equation \ref{eq6}.
        \ENDFOR
        \STATE Generate fused features $\mathbf{h}(\mathbf{x})$ with spatial weighting using Equation \ref{eq7}.
        \STATE Predict corresponding physical quantity $\hat{\mathbf{u}}(\mathbf{x})$ with decoding head using Equation \ref{eq9}.
    \ENDFOR
    \STATE Compute PDE residual loss and boundary/initial condition loss, then calculate the total weighted loss using Equation \ref{eq4}.
    \STATE Update parameters $\{\mathcal{Z}, \theta_s, \psi\}$ via gradient descent algorithms (e.g., Adam).
\ENDWHILE
\end{algorithmic}
\end{algorithm}

\subsection{Physical Decoding}
    The Continuous Feature Synthesis stage yields a latent representation $\mathbf{h}(\mathbf{x}) \in \mathbb{R}^H$, which encodes the rich localized spatiotemporal dynamics but remains in an abstract feature space. To map these abstract features to the target physical quantities (e.g., velocity field, pressure), a lightweight physical decoding head is employed, denoted as $\phi_{\psi}: \mathbb{R}^H \to \mathbb{R}^{D_{out}}$. This is typically implemented as a shallow MLP or a linear projection layer. The final predicted solution at any coordinate $\mathbf{x}$ is obtained by:
    \begin{equation}
    \label{eq9}
        \hat{\mathbf{u}}(\mathbf{x}) = \phi_{\psi}\big(\mathbf{h}(\mathbf{x})\big),
    \end{equation}
    where $\hat{\mathbf{u}}(\mathbf{x})$ represents the approximated physical fields. Then, the complete trainable parameters $\Theta=\{\mathcal{Z},\theta_s,\psi\}$ are updated via gradient descent by minimizing Equation \ref{eq4}. 
    
    The overall training procedure of our proposed framework is summarized in Algorithm \ref{alg:method}. It illustrates how the discrete latent contexts and the continuous neural decoder are jointly optimized under the constraints of physics-informed residuals. Note that since both the synthesis operator $f_{\theta_s}$ and the decoding head $\phi_{\psi}$ are constructed using smooth activation functions (e.g., Tanh), the constructed solution $\hat{\mathbf{u}}(\mathbf{x})$ is guaranteed to be $C^\infty$ continuous w.r.t. the input coordinates $\mathbf{x}$. Consequently, high-order derivatives (e.g., $\nabla^2 \hat{\mathbf{u}}$) can be computed exactly via automatic differentiation, allowing the entire architecture to be trained end-to-end.

\section{Experiments}
    To validate the effectiveness of PILIR in mitigating spectral bias, we perform numerical experiments on a range of challenging problems. These include oscillatory Helmholtz equations in both 2D and 3D to assess the method's capability in handling globally high-frequency solutions, as well as multiscale Allen–Cahn, convection, reaction–diffusion, and 3D Navier–Stokes equations to evaluate its performance in capturing fine-grained details. To compute the accuracy of the approximated solutions, the relative L2 error is used, defined as $\frac{||\mathbf{u}-\hat{\mathbf{u}}||_2}{||\mathbf{u}||_2}$. The experimental results are obtained at seed values of 100, 200, 300, 400, and 500, respectively.

    We compare our method against the following approaches: (1) The baseline method PINN~\cite{raissi2019physics}; (2) PINN‑Wavelet~\cite{uddin2023wavelets}, which uses a Gaussian wavelet activation function to enrich the frequency of latent features, defined as $\sigma(x) = -x e^{-\frac{1}{2}x^{2}}$; (3) PIXEL~\cite{kang2023pixel}, a grid-based approach which uses multi‑grid representation with grid offset technique; and (4) PIG~\cite{kang2025pig}, also a grid-based approach, which uses an adaptive mesh technique and Gauss kernel interpolation. Training details and PDE formulations are provided in the Appendix A.3, and extra comparisons are included in the Appendix A.6.

\begin{table*}[htbp]
\centering
    \begin{tabular*}{\textwidth}{@{\extracolsep{\fill}}lrrrrrr@{}}
        \toprule
        \textbf{Methods} & \textbf{Helmholtz-3D} & \textbf{Helmholtz-2D} & \textbf{Convection} & \textbf{MS-Convection} & \textbf{Allen-Cahn} & \textbf{Reaction-Diffusion} \\
        \midrule
        PINN  & 2.47e+00 & 1.62e+00  & 1.32e-02  &3.65e-01 & 5.21e-01  & 4.56e-02  \\
              & $\pm$ 2.92e+00 & $\pm$ 1.25e+00 & $\pm$ 5.38e-03 &$\pm$ 1.04e-02 & $\pm$ 7.68e-03 & $\pm$ 1.23e-03 \\
        \midrule
        Wavelet  & 1.05e+00 & 4.96e-01 & 1.31e-02 &4.09e-01 & 3.47e-01 & 4.37e-02  \\
                & $\pm$ 8.99e-02 & $\pm$ 2.74e-01 & $\pm$ 9.36e-03 &$\pm$ 1.60e-02 & $\pm$ 1.69e-01 & $\pm$ 2.36e-03 \\
        \midrule
        PIXEL & 1.36e+01  & 1.63e+00 	& 4.59e-02  &4.54e-01 & 1.02e-01  & 2.38e-01  \\
               & $\pm$ 4.27e+00 & $\pm$ 7.55e-01 & $\pm$ 2.83e-03 &$\pm$ 2.98e-03 & $\pm$ 2.88e-02 & $\pm$ 4.97e-03 \\
        \midrule
        PIG  & OOM & 1.00e+00  & 5.90e-03   &3.77e-01  & \textbf{1.48e-03}  & 5.66e-02  \\ 
             & &$\pm$ 2.81e-03 &$\pm$ 3.29e-03 &$\pm$ 1.32e-03 &$\pm$ 7.15e-04 &$\pm$ 1.44e-02 \\
        \midrule
        PILIR (ours) & \textbf{5.04e-02} & \textbf{1.36e-02} & \textbf{2.78e-03} &\textbf{1.46e-01}  & 3.26e-02 & \textbf{1.90e-02} \\
                      & $\pm$ 1.24e-02 & $\pm$ 4.21e-03 & $\pm$ 9.33e-04 &$\pm$ 2.70e-02 & $\pm$ 2.60e-03 & $\pm$ 1.35e-02 \\
        \bottomrule
    \end{tabular*}
\caption{Comparison results of PILIR with state-of-the-art algorithm on the forward problem. Mean and std Values are reported.}
\label{tab1}
\end{table*}

\begin{figure*}[htbp]
\centering
\includegraphics[width=1\linewidth]{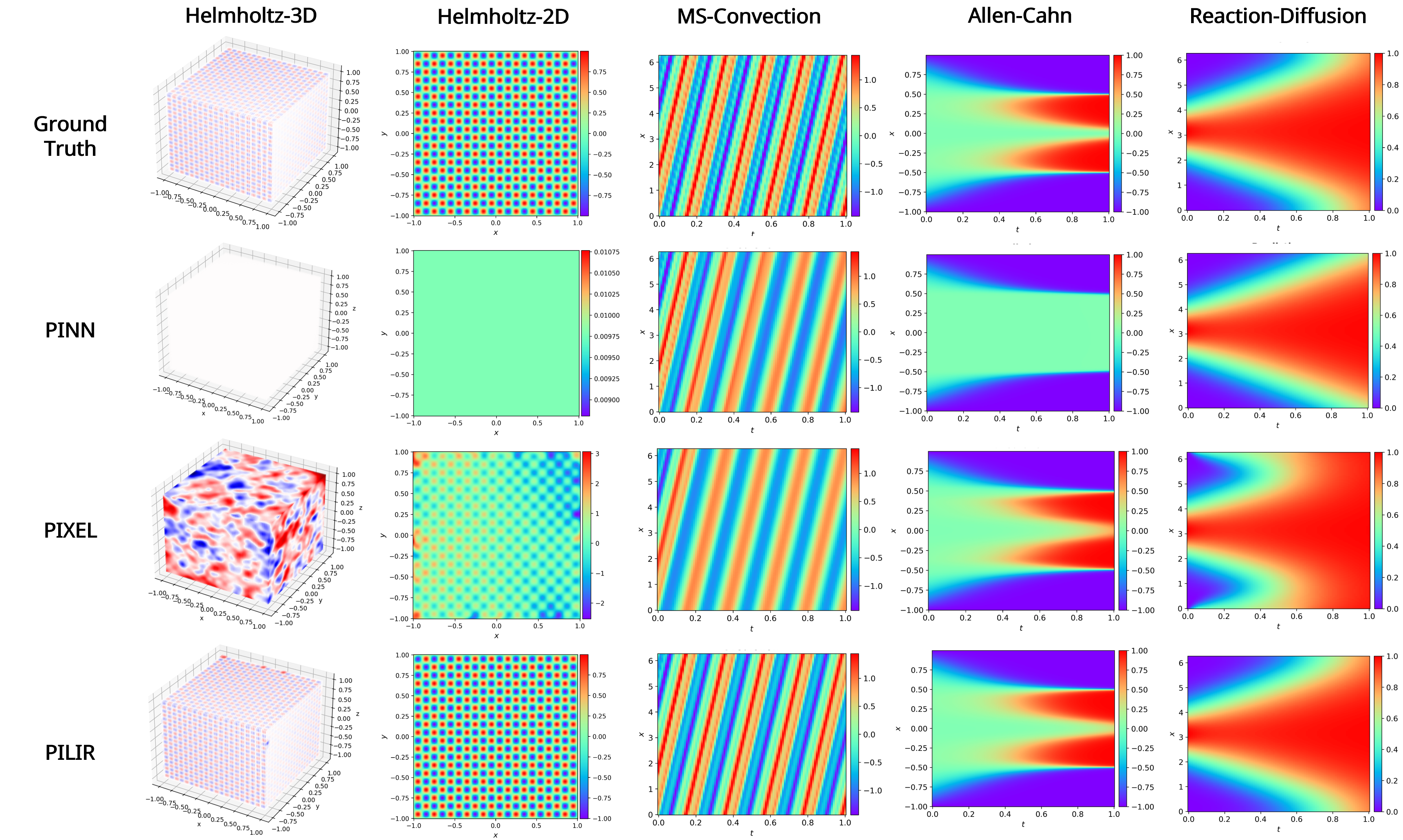}
\caption{Figure results for Ground Truth, PINN, PIXEL, and PILIR.}
\label{fig2}
\end{figure*}

\subsection{Results and Discussion}

\begin{figure*}[htbp]
    \centering
    \includegraphics[width=1\linewidth]{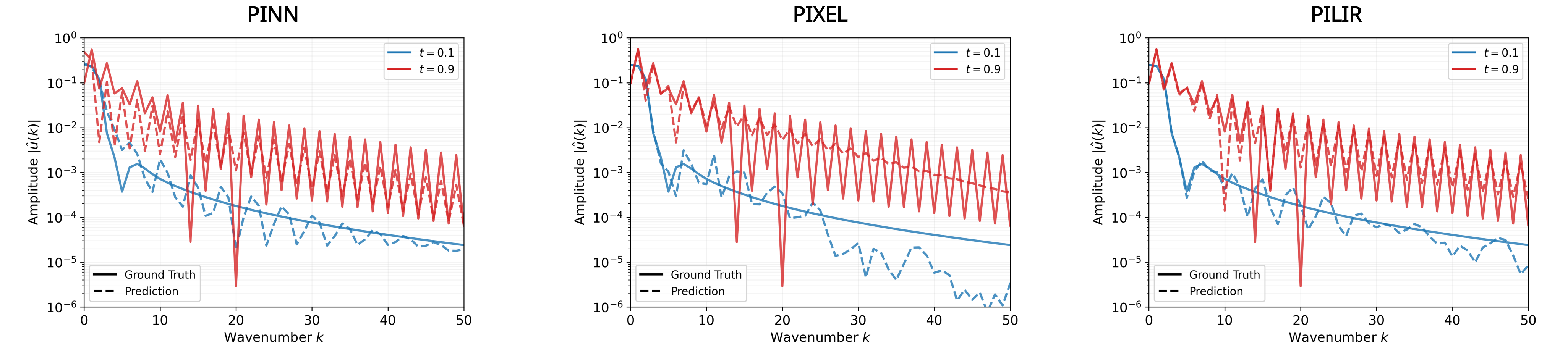}
    \caption{Spectrum analysis on Allen-Cahn equation.}
    \label{fig3}
\end{figure*}

\subsubsection{Forward Problem}
    In the forward problem, a neural network serves as a surrogate model to approximate the exact solution at given query coordinates. Quantitative results are presented in Table \ref{tab1}, and corresponding visualizations are presented in Figure \ref{fig2}. Please refer to Appendix A.5 for extra visualizations.

\begin{figure*}[htbp]
    \centering
    \includegraphics[width=1\linewidth]{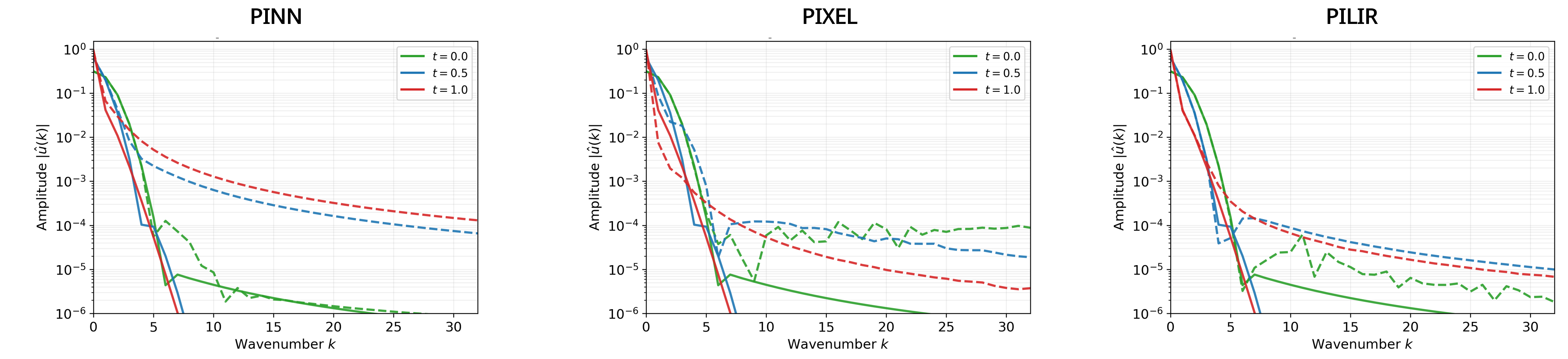}
    \caption{Spectrum analysis on Reaction-Diffusion equation.}
    \label{fig4}
\end{figure*}  

    In both Helmholtz equations, we use high-frequency parameters $a_1=10, a_2=10$ for 2D, and an extra parameter $a_3=10$ for 3D, resulting in a very oscillatory solution. As expected, PINN fails to converge to an accurate solution due to the spectral bias. Although PIXEL generates more similar patterns, the positions between wave peaks may not be recoverable due to the Nyquist sampling theorem without careful adjustment of the grid resolution, which highlights the limitations of traditional interpolation functions. Wavelet activation functions can provide some high-frequency information, but they still fall short of our approach using generative features. 
    
    It is worth noting that, although the adaptive mesh technique proposed by PIG alleviates the need for increased mesh resolution in complex multiscale problems, it requires storing additional positional information and employs Gaussian kernels to aggregate features across all grid points, thereby demanding greater computational resources. We observed memory exhaustion in 3D problems, therefore we mark these cases as OOM (Out of Memory). Additionally, regarding the problem of global oscillations or wave propagation, we hypothesize that the positions of grid points continuously shift and fail to converge to the region that is most difficult to fit, resulting in suboptimal performance.

    For the Allen-Cahn equation, previous studies have shown that PINNs perform poorly on this notoriously difficult problem without employing additional training techniques, such as causal learning~\cite{wang2024respecting}. In this case, the wavelet activation function can only improve PINN's performance to a limited extent due to its global applicability. On the contrary, Grid-based PINNs and our PILIR can focus on local information with greater precision, enabling more accurate fitting in complex bottleneck areas (e.g., the rapidly evolving solutions at $t=1$). While PIG performs best in this equation, it suffers from a larger computation resource requirement and time cost, as mentioned before. Figure \ref{fig3} presents the spectrum analysis using Fast Fourier Transform (FFT). We can clearly see that at $t=0.9$, PINN fails to learn the top-10 frequency components with the largest amplitude. While PIXEL successfully captures this part, it does not reduce the amplitude of high-frequency components and generates a relatively smooth curve, which may result in blurring (e.g., artifacts) at the boundary. 

    For the convection and reaction-diffusion equations, PILIR still made considerable improvements. Figure \ref{fig4} presents the spectrum analysis of the reaction-diffusion equation, from which we can see that the diffusion phenomenon causes the high-frequency components to gradually fade away. As presented, PINN maximally preserves the high-frequency components after $t>0.5$, and PIXEL generates severe false high-frequency information at $t=0$. Please refer to Appendix A.4 for more spectrum analysis.

\begin{table}[tb]
\centering
    \begin{tabular}{lrrrr}
        \toprule
        \textbf{Methods} & $u$ & $v$  & $\lambda_1$ & $\lambda_2$ \\
        \midrule
        PINN & 6.43e-03  & 2.00e-02  & 9.99e-01  & 1.07e-02  \\
        \midrule
        Wavelet & 4.46e-03 	& 1.43e-02  & 9.99e-01  & \textbf{1.06e-02}  \\
        \midrule
        PIXEL & 6.63e-03 	& 2.37e-02  & 9.56e-01  & 1.45e-02  \\
        \midrule
        PILIR & \textbf{2.58e-03} & \textbf{9.72e-03}  & \textbf{1.00e+00} & \textbf{1.06e-02} \\
        \bottomrule
    \end{tabular}
\caption{Comparison results of PILIR with state-of-the-art algorithm on Navier-Stokes equation. Mean values are reported.}
\label{tab2}
\end{table}

\begin{table}[tb]
\centering
    \begin{tabular}{lrrr}
        \toprule
        \textbf{Methods} & Allen-Cahn & Helmholtz & Convection \\
        \midrule
        PIXEL $8\times8$ & 3.05e-01   & 6.41e+00  & 5.10e-01  \\
        PILIR $8\times8$ & 3.95e-02   & 2.25e-01  & 1.25e-02  \\
        \midrule
        PIXEL $12\times12$ & 1.41e-01  & 2.02e+00  & 7.64e-02  \\
        PILIR $12\times12$ & 3.39e-02  & 1.29e-01  & 2.77e-03  \\
        \midrule
        PIXEL $16\times16$ & 1.02e-01  & 1.63e+00  & 4.59e-02  \\
        PILIR $16\times16$ & 3.26e-02  & 1.36e-02  & 2.78e-03  \\
        \bottomrule
    \end{tabular}
\caption{Comparison results of PILIR under different grid resolution. Mean values are reported.}
\label{tab3}
\end{table}

\begin{figure*}[htb]
    \centering
    \includegraphics[width=1\linewidth]{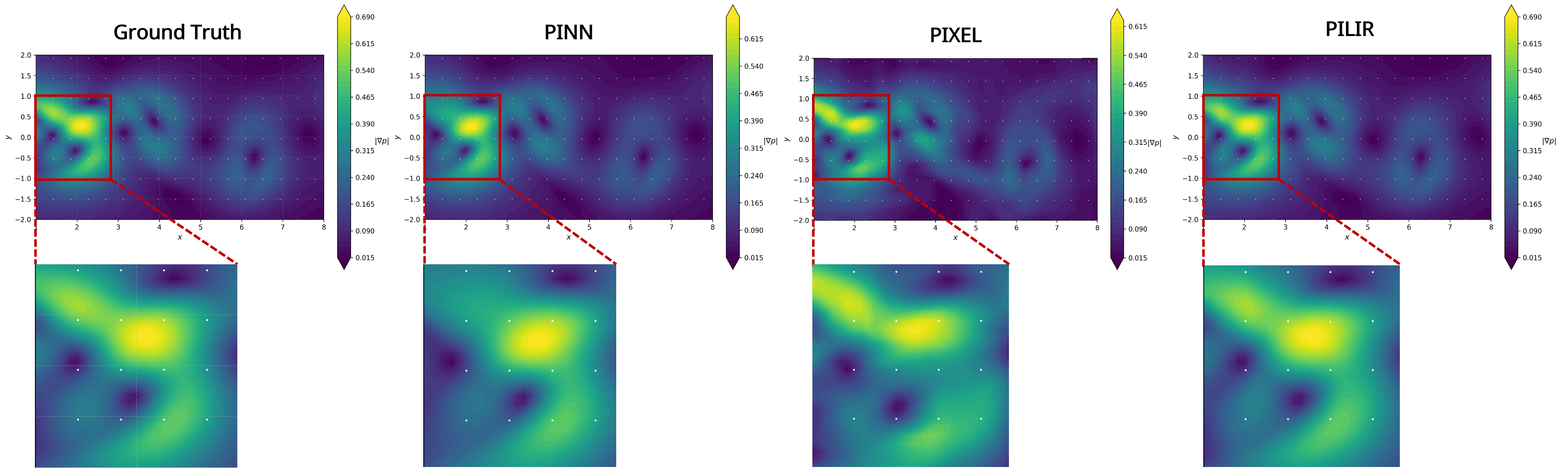}
    \caption{Comparison of pressure gradients on Navier-Stokes equation.}
    \label{fig5}
\end{figure*}

\begin{figure*}[htb]
    \centering
    \includegraphics[width=1\linewidth]{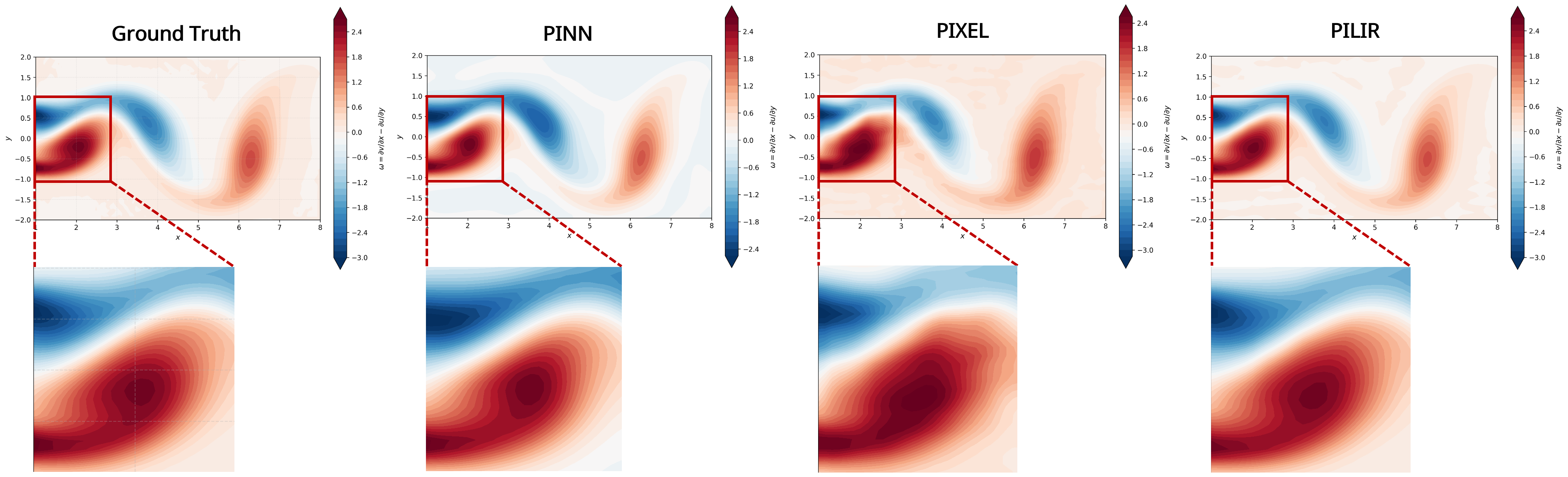}
    \caption{Comparison of vorticity on Navier-Stokes equation.}
    \label{fig6}
\end{figure*}

\subsubsection{Inverse Problem}
    Inverse problem refers to solving PDEs and determining their unknown coefficients using existing data. We consider the most challenging incompressible Navier-Stokes equation with $\lambda_1=1,\lambda_2=0.01$, which requires the network to produce the speed $u,v$ and pressure $p$ simultaneously. Quantitative results are presented in Table \ref{tab2}, and we omit PIG due to limited GPU memory.

    As shown in the table, PILIR accurately reconstructs both the solution fields and the unknown equation coefficients. The pressure field $p$ is omitted from the results, as the training data contain only velocity components $u$ and $v$, and the governing equation includes $p$ only within spatial derivatives. Consequently, the pressure predicted by the network may exhibit numerical offsets from the reference solution.

    To ensure a precise comparison, we compute the absolute pressure gradient at the final time step as 
    \( |\nabla p| = \sqrt{(\partial_x p)^2 + (\partial_y p)^2} \). The results in Figure~\ref{fig5} demonstrate that PILIR recovers finer structural details more accurately.
    
    To further examine whether the network has genuinely learned the underlying physics or merely memorized the data distribution, we also visualize the vorticity field, defined by $\omega = \frac{\partial v}{\partial x} - \frac{\partial u}{\partial y}$, which directly reflects the learned velocity gradients. As shown in Figure~\ref{fig6}, although PINN achieves 
    a lower L2 error on the velocity field compared to PIXEL, it fails to capture the correct vorticity patterns. In contrast, PILIR reproduces the vorticity field with significantly finer and more accurate details.

\subsection{Ablation Study}
    To validate the effectiveness of our generative decoding mechanism, we perform a sensitivity analysis with respect to grid resolution. A key premise of our approach is that it decouples reconstruction quality from the underlying grid density, enabling the neural operator to synthesize sub‑grid details that cannot be recovered by purely deterministic interpolation. As summarized in Table~\ref{tab3}, while the overall accuracy declines at lower resolutions, the drop remains modest. Even at an $8\times8$ resolution, our method still surpasses PIXEL with $16\times16\times16$ multi‑grid representations, highlighting the advantage of learnable feature fusion over deterministic interpolation. Especially for the high-frequency 2D Helmholtz problem, PILIR can achieve an effective result using only 64 grid points.

\section{Conclusion}
    This paper introduces PILIR, a hybrid paradigm that bridges discrete grid encoding with continuous neural reconstruction to overcome the spectral bias in physics-informed modeling. Drawing inspiration from local implicit image functions, our approach treats grid vertices as carriers of latent local context rather than scalar intensities. This design empowers a neural operator to synthesize sub-grid details, effectively recovering high-frequency dynamics that standard methods often miss. Experimental results across diverse multi-scale PDEs demonstrate that PILIR achieves fidelity superior to existing methods. Notably, the proposed method maintains robustness even on coarse grids by decoupling reconstruction quality from grid resolution through a learnable feature fusion mechanism. Consequently, this hybrid discrete-continuous strategy provides an effective and scalable solution for high-fidelity physics modeling.
\bibliographystyle{named}
\bibliography{ijcai26}

\clearpage
\appendix
\section{Appendix}

% \subsection{Code availability}
%     The code for PILIR will be made publicly available upon acceptance of the paper.

% \subsection{LLM Statement}
%     This paper used Gemini 3 pro preview to polish the language and improve the clarity of the writing. The authors are responsible for all content in this paper.

\subsection{PDE Details and Experimental Setup}
    All experiments are trained for 100,000 epochs using the Adam optimizer. A cosine annealing schedule is adopted for the learning rate, with a lower bound of \(1\times10^{-6}\). For 2D problems, we sample 10,000 collocation points for the PDE residual and 1,000 points for the initial and boundary conditions. For 3D problems, the number of sample points is increased by a factor of five. To evaluate the robustness of different methods under gradient conflicts, we use a simple loss‑weighting scheme: \(\lambda_{\text{ic}} = \lambda_{\text{bc}} = \lambda_{\text{r}} = 1\).

    The decoding head in PIXEL, PIG and PILIR is implemented as a shallow MLP with one hidden layer of 16 neurons, using the hyperbolic tangent (tanh) activation function. The latent feature dimension of each grid point is fixed at 4. In all 2D experiments, PIXEL employs a fixed grid resolution of \(16\times16\times16\) (where the first dimension denotes multiple grids). PILIR uses a grid resolution of \(1\times16\times16\), except for the 1D convection and Allen‑Cahn equations, where a resolution of \(16\times16\times16\) is adopted. For solving 3D problems, a z‑axis with a resolution of 16 is added to the grid. For PIG, the initial sigma is fixed at 0.05. The feature fusion network in PILIR has a hidden dimension of 16 and 2 hidden layers, except for the 1D convection and Allen‑Cahn equations, where 3 hidden layers are used. All models are trained on a single NVIDIA RTX 5090 GPU with 32 GB of memory.
    
\subsubsection{Helmholtz Equation}
    The 3D Helmholtz equation is defined as:
    \begin{align}
        \Delta u + k^2 u &= q, \quad (x,y,z) \in [-1,1]^3, \\
        u &= 0, \quad (x,y,z) \in \partial[-1,1]^3,
    \end{align}
    where $\Delta = \partial_{xx} + \partial_{yy} + \partial_{zz}$ is the Laplacian operator, and $k$ is the wavenumber. The source term $q$ is given by:
    \begin{equation}
        \begin{aligned}
            &q(x,y,z) = k^2\sin(a_1\pi x) \sin(a_2\pi y) \sin(a_3\pi z)\\
            &-\pi^2(a_1^2 + a_2^2 + a_3^2)\sin(a_1\pi x) \sin(a_2\pi y) \sin(a_3\pi z),
        \end{aligned}
    \end{equation}
    which yields the exact solution:
    \begin{align}
        u(x,y,z) = \sin(a_1\pi x) \sin(a_2\pi y) \sin(a_3\pi z).
    \end{align}
    In our experiments, we set $a_1=10$, $a_2=10$, $a_3=10$, and $k=1$. For the 2D Helmholtz equation, the $z$‑dependence is removed, resulting in the exact solution $u(x,y) = \sin(a_1\pi x) \sin(a_2\pi y)$. The initial learning rate is set to 0.01. For the PINN model, we use 7 hidden layers and 100 hidden dimensions. For PIG, the number of Gaussians is set to 1200.
    
\subsubsection{Convection Equation}
    The 1D convection equation is given by
    \begin{align}
        u_t + \beta u_x = 0, \qquad x \in [0,2\pi],\; t \in [0,1],
    \end{align}
    with \(\beta = 30\). Periodic boundary condition $u(0,t)=u(2\pi,t)$ is imposed. Two initial conditions are used:
    \begin{itemize}
        \item $u(x,0) = \sin x + 0.5\sin(4x) + 0.1\sin(8x) + 0.1\sin(16x)$ for comparing methods on a challenging multi‑scale problem;
        \item $u(x,0) = \sin x$ for the ablation study on grid‑resolution sensitivity in PILIR.
    \end{itemize}
    The initial learning rate is set to 0.001. For the PINN model, we use 3 hidden layers with 50 neurons each. For PIG, the number of Gaussians is set to 800.

\subsubsection{Allen-Cahn Equation}
    The Allen-Cahn equation is given by:
    \begin{align}
        u_t - \nu u_{xx} + \lambda u^3 - \lambda u &= 0, \quad x \in [-1,1], \; t \in [0,1],
    \end{align}
    with $\nu = 0.0001$ and $\lambda = 5$. Periodic boundary conditions are imposed:
    \begin{align}
        u(-1,t) &= u(1,t), \\
        u_x(-1,t) &= u_x(1,t).
    \end{align}
    The initial condition is set to:
    \begin{align}
        u(x,0) = x^2 \cos(\pi x).
    \end{align}
    The initial learning rate is set to 0.001. For the PINN model, we use 6 hidden layers and 128 hidden dimensions. For PIG, the number of Gaussians is set to 800.

\subsubsection{Reaction-Diffusion Equation}
    The reaction-diffusion equation is given by
    \begin{align}
        u_t - \nu u_{xx} - \rho u(1-u) = 0, \qquad x \in [0,2\pi], \; t \in [0,1],
    \end{align}
    with $\nu=0.5$ and $\rho=5$. Periodic boundary conditions $u(0,t)=u(2\pi,t)$ are imposed. The initial condition is set to:
    \begin{equation}
        e^{-\frac{1}{2}\bigl(\frac{4(x-\pi)}{\pi}\bigr)^2}.
    \end{equation}
    The initial learning rate is set to 0.001. For the PINN model, we use 3 hidden layers with 50 neurons each. For PIG, the number of Gaussians is set to 400. For PILIR, the latent feature dimension of each grid point is set to 8.

\subsubsection{Navier-Stokes Equation}
    The 3D incompressible Navier-Stokes equations is given by:
    \begin{equation}
        \begin{aligned}
            u_t+\lambda_1(uu_x+vu_y) &= -p_x+\lambda_2(u_xx+v_yy), \\
            v_t+\lambda_1(uv_x+vv_y) &= -p_y+\lambda_2(v_xx+v_yy), \\
            u_x + v_y &= 0, \\
        \end{aligned}
    \end{equation}
    where $(x,y) \in [1,8] \times [-2,2]$ and $t \in [0,20]$, with $\lambda_1 = 1$ and $\lambda_2 = 0.01$. The initial learning rate is set to 0.01. For the PINN model, we use 9 hidden layers and 20 hidden dimensions.

\begin{table*}[htbp]
\centering
    \begin{tabular*}{\textwidth}{@{\extracolsep{\fill}}lrrrrrr@{}}
        \toprule
        \textbf{Methods} & \textbf{Helmholtz-3D} & \textbf{Helmholtz-2D} & \textbf{Convection} & \textbf{MS-Convection} & \textbf{Allen-Cahn} & \textbf{Reaction-Diffusion} \\
        \midrule
        MSPINN & 8.27e+00 & 5.64e+00  & 7.06e-03  & 4.64e-01 & 2.96e-02  &  2.99e-01 \\
        \midrule
        PIKAN & OOM & 1.00e+00  & 9.19e-04  & 8.98e-02 & 5.63e-03  & 2.24e-02  \\
        \midrule
        PILIR & 3.12e-02 & 8.83e-03 & 1.86e-03 & 1.12e-01  & 2.94e-02 & 9.50e-03 \\
        \bottomrule
    \end{tabular*}
\caption{Comparison results of PILIR, MSPINN and PIKAN. Best values are reported.}
\label{tab4}
\end{table*}

\begin{figure*}[htb]
    \centering
    \includegraphics[width=1\linewidth]{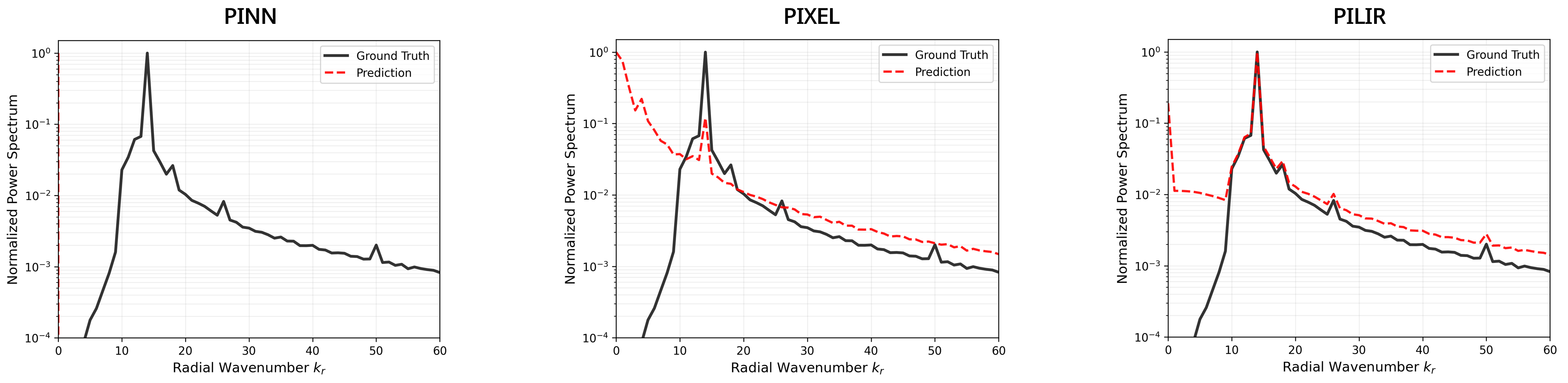}
    \caption{Spectrum analysis on Helmholtz equation.}
    \label{fig7}
\end{figure*}

\begin{figure*}[htb]
    \centering
    \includegraphics[width=1\linewidth]{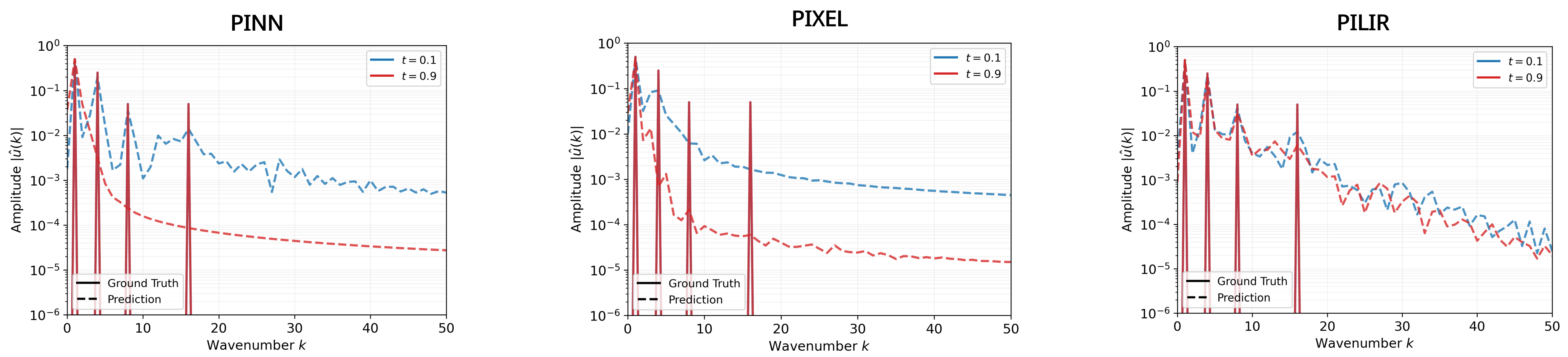}
    \caption{Spectrum analysis on multi-scale convection equation.}
    \label{fig8}
\end{figure*}

\subsection{Extra Spectrum Analysis}
\subsubsection{2D Helmholtz Equation}
    The spectral analysis for the 2D Helmholtz equation is shown in Figure~\ref{fig7}. It can be seen that the standard PINN fails to resolve high‑frequency content, recovering only the DC component ($k_r=0$) and consequently yielding a nearly zero output. Both PIXEL and PILIR are able to learn high‑frequency modes, but PILIR exhibits superior accuracy in the dominant high‑frequency region ($k_r\approx 10$).

\subsubsection{Multi‑scale Convection Equation}
    Figure~\ref{fig8} displays the spectral analysis of the multi‑scale convection equation. PILIR accurately captures frequency components of different scales and amplitudes. Convection problems are characterized by wave‑like propagation where the wave energy should remain invariant, hence the solver must be free of numerical dissipation. We can see that only PILIR maintains a faithful representation of the multi‑scale frequency details at the later time $t=0.9$.

\subsubsection{Navier–Stokes Equation}
\begin{figure*}[h]
    \centering
    \includegraphics[width=1\linewidth]{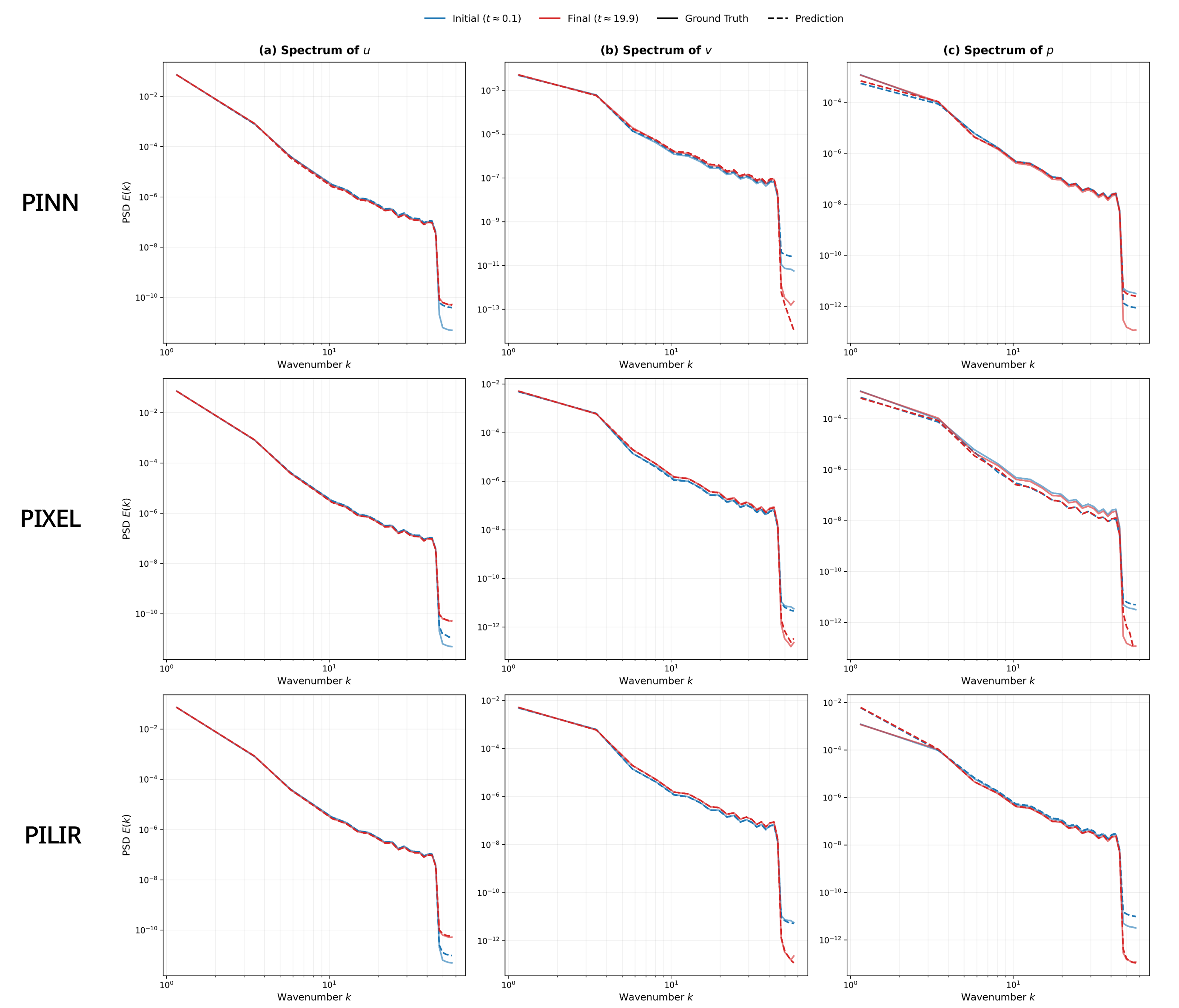}
    \caption{Spectrum analysis on Navier-Stokes equation.}
    \label{fig9}
\end{figure*}

    The spectral analysis for the Navier–Stokes equation is provided in Figure~\ref{fig9}. Among the compared methods, PILIR most faithfully reproduces the frequency content of the horizontal velocity, vertical velocity, and pressure fields.

% \subsection{Extra Visualization}
%     From Figure~\ref{fig10} to Figure~\ref{fig11} we show the ground‑truth solutions, the predictions of our model, and the corresponding pointwise absolute errors for each two‑dimensional test case. For time‑dependent PDEs we also include the solution profiles at several representative time slices.

\subsection{Comparison with PIKAN and MSPINN}
    We further compare PILIR with two alternative approaches: MSPINN~\cite{wang2021eigenvector}, which employs multi‑scale spatial‑temporal Fourier embeddings, and PIKAN~\cite{wang2025kolmogorov}, which replaces the standard MLP with a Kolmogorov–Arnold Network (KAN) architecture. Quantitative results are summarized in Table~\ref{tab4}. 

    For MSPINN, choosing an appropriate set of frequency scales is non‑trivial. We set them heuristically based on the visual inspection of the problem spectrum: spatial scales [1, 5] for reaction‑diffusion, [1] for convection, [1, 10, 20] for multi‑scale convection, [1, 10, 50] for Allen‑Cahn, and [1, 10] for the Helmholtz equation. Temporal scales are fixed at [1] when the PDE is time‑dependent. Although MSPINN improves upon the vanilla PINN on the Allen‑Cahn and convection problems, it does not perform consistently well across the other cases.

    PIKAN shows strong performance on multi‑scale problems, especially when high‑frequency modes are non‑dissipative, as in the Allen‑Cahn and convection equations. However, it is less effective than PILIR on globally high‑frequency problems and on problems where high‑frequency components decay rapidly, such as the reaction‑diffusion equation. In our experiments we keep the hidden‑layer dimensions and network depth the same as in the baseline PINN. Nevertheless, the repeated tuning of the B‑spline parameters in KAN leads to a substantial increase in both training time and memory consumption, which is roughly eight times higher than that of PILIR.

    It is worth noting that PIKAN and PILIR represent largely orthogonal strategies. The former modifies the network architecture, while the latter focuses on advanced input encoding. Exploring ways to combine their strengths, for instance, by partially replacing PILIR’s backbone with a KAN‑like structure could be a promising direction to improve adaptability across diverse problem types and potentially reduce memory overhead. Such an investigation, however, lies beyond the scope of the present study.

\clearpage

\end{document}